\documentclass[conference]{IEEEtran}
\IEEEoverridecommandlockouts
\usepackage[utf8]{inputenc}
\usepackage[T1]{fontenc}

\usepackage{amsmath, amssymb, amsfonts}

\usepackage{algorithm}   
\usepackage{algpseudocode}

\usepackage{graphicx}
\usepackage{svg}
\usepackage{wrapfig}
\usepackage{subcaption}
\usepackage{placeins} 
\usepackage{stfloats}   
\usepackage{float}

\usepackage{booktabs}     
\usepackage{multirow}
\usepackage{makecell}
\usepackage{adjustbox}
\usepackage{array}
\usepackage[table]{xcolor} 

\usepackage{textcomp}
\usepackage{nicefrac}
\usepackage{microtype}

\usepackage{hyperref}
\usepackage{url}

\usepackage{cite}

\usepackage{caption}
\def\BibTeX{{\rm B\kern-.05em{\sc i\kern-.025em b}\kern-.08em
    T\kern-.1667em\lower.7ex\hbox{E}\kern-.125emX}}
\begin{document}

\title{Highly Imbalanced Regression with \\ Tabular Data in SEP and Other Applications}


\author{
\IEEEauthorblockN{Josias K.~Moukpe\IEEEauthorrefmark{1}, Philip K.~Chan\IEEEauthorrefmark{1}, Ming Zhang\IEEEauthorrefmark{2}}
\IEEEauthorblockA{\IEEEauthorrefmark{1}Department of Electrical Engineering and Computer Science\\
\IEEEauthorrefmark{2}Department of Aerospace, Physics and Space Sciences\\
Florida Institute of Technology, Melbourne, FL, USA\\
jmoukpe2016@my.fit.edu, \{pkc, mzhang\}@fit.edu}
}

\maketitle

\begin{abstract}
We investigate imbalanced regression with tabular data that have an imbalance ratio larger than 1,000 ("highly imbalanced").  Accurately estimating the target values of rare instances is important in applications such as forecasting the intensity of rare harmful Solar Energetic Particle (SEP) events.
For regression, the MSE loss does not consider the correlation between predicted and actual values.  Typical inverse importance functions allow only convex functions.  Uniform sampling might yield mini-batches that do not have rare instances.
We propose CISIR that incorporates correlation, Monotonically Decreasing Involution (MDI) importance, and stratified sampling. Based on five datasets, our experimental results indicate that CISIR can achieve lower error and higher correlation than some recent methods. Also, adding our correlation component to other recent methods can improve their performance. Lastly, MDI importance can outperform other importance functions. Our code can be found in \href{https://github.com/Machine-Earning/CISIR}{https://github.com/Machine-Earning/CISIR}.
\end{abstract}


\section{Introduction}
\label{sec:intro}

Recent work on imbalanced regression usually focuses on images \cite{yang2021ldsfds,ren2022bmse,gong2022ranksim,keramati2023conr,xiong2024hca}.  
In this study, we investigate imbalanced regression with tabular data that have an imbalance ratio larger than 1,000 ("highly imbalanced").  We investigate methods without substantive additional computation due to factors such as additional synthetic data and pre-training for representation learning.

Accurately estimating the target values for rare instances is important in various applications.  For example, forecasting the intensity of rare Solar Energetic Particle (SEP) \cite{reames2021solar}  events helps provide warnings to alert astronauts to seek shelter and protect sensitive equipment. In this study, one SEP application is to forecast the future proton intensity based on features from electron intensity \cite{posner2007up} and Coronal Mass Ejections (CMEs).  
Another is to forecast the peak proton intensity based on features from CMEs \cite{richardson2018prediction}.  Three other applications are large torque values for a robot arm, high popularity in online news, and large amounts of feedback for blogs.

\begin{figure}[!t] 
    \centering
    \includesvg[width=0.6\columnwidth]{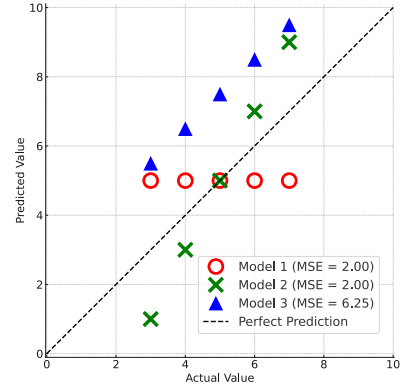}
    \caption{An issue with MSE.}
    \label{fig:threemodels}
\end{figure}

For regression, the typical loss function used for training a neural network model is the Mean Squared Error (MSE) between the model predictions and the actual target labels. 
However, MSE does not consider the correlation between predicted values and actual values.  Consider evaluating 3 models on 5 instances in Figure~\ref{fig:threemodels}.  Model 1 (red) and Model 2 (green) have the same MSE of 2.  According to MSE, Model 1 and Model 2 are equally inaccurate.  However, the (same) predicted values of Model 1 are not correlated with the actual values, while the predicted values of Model 2 are positively correlated with the actual values.  The predicted values should be positively correlated with the actual values, similar to the perfect reference (dotted black).  Also, perfect MSE yields perfect positive correlation, while perfect positive correlation does not guarantee perfect MSE. For example, Model 3 (blue) in Figure~\ref{fig:threemodels} has a perfect positive correlation but a larger MSE than the other two models. Hence, while MSE alone is not sufficient, correlation alone is also not sufficient. Particularly, for rare and important instances in the real world (e.g. harmful SEP events), when the predicted values are increasing, we would like the actual values to be increasing correspondingly, which implies positive correlation. Otherwise, without correlation, the actual value might be increasing or decreasing, which implies more uncertainty. Hence, for imbalanced regression, we propose a weighted version of Pearson Correlation Coefficient (wPCC) to supplement the weighted MSE.

To handle imbalanced regression, one method is to weight rarer instances more in the loss function.   Based on the probability density, the weight for each instance is governed by what we call an "importance function."  We use "importance" to distinguish it from a "weight" in the model.  
One typical importance function is the inverse function \cite{yang2021ldsfds} that aims to achieve a balanced distribution.  
DenseLoss \cite{steininger2021density} uses a linear function.  Generally, we do not know the appropriate importance function for an application because it
is dependent on the dataset, imbalance ratio, and loss function. Hence, we propose Monotonically Decreasing Involution (MDI) importance, which allows a family of convex, linear, and concave functions.  Also, MDI importance preserves the involution property exhibited by the inverse function that achieves the balanced distribution.

For highly imbalanced datasets, some of the mini-batches used in stochastic gradient descent (SGD) via uniform random sampling might not have any rare instances.  That is, for some model updates, the gradient does not consider rare instances.  To ensure rare instances are represented in each mini-batch, we propose stratified sampling for the mini-batches.

Our overall approach is called CISIR (Correlation, Involution importance, and Stratified sampling for Imbalanced Regression).
Our contributions include:
\begin{itemize}
    \item proposing weighted Pearson Correlation Coefficient (wPCC) as a secondary loss function,
    \item introducing MDI importance that allows convex, linear, and concave functions,
    \item stratified sampling in mini-batches to represent rare instances in each mini-batch,
    \item CISIR can achieve lower error and higher correlation than other recent methods,
    \item adding wPCC to other methods is beneficial in not only increasing correlation, but also reducing error, and
    \item MDI importance can outperform other importance functions.
\end{itemize}

\section{Related Work}
\label{sec:related}

Recent methods for imbalanced regression can be clustered into four groups: distribution resampling, label-space smoothing, representation-space calibration, and loss re-weighting.
Distribution re-sampling methods directly modify the training data distribution by generating synthetic samples for rare label ranges. SMOGN~\cite{branco2017smogn} and SMOTEBoost-R~\cite{moniz2018smoteboost} adapt popular synthetic oversampling techniques originally developed for classification tasks to continuous targets.

Label-space smoothing approaches adjust the label distribution or its granularity to alleviate the imbalance. LDS~\cite{yang2021ldsfds} smooths the empirical label density via Gaussian kernel convolution, enabling re-weighting schemes based on a more robust label distribution. HCA~\cite{xiong2024hca} constructs a hierarchy of discretized labels at varying granularities, using predictions from coarser levels to refine fine-grained predictions, balancing quantization errors and prediction accuracy.

Representation-space calibration tackles imbalance by enforcing structural regularities directly in latent feature space. FDS~\cite{yang2021ldsfds} aligns latent representations with smoothed label distributions. RankSim~\cite{gong2022ranksim} explicitly calibrates the representation space to reflect the pairwise ranking structure in labels. ConR~\cite{keramati2023conr} introduces a contrastive regularizer that penalizes incorrect proximities in feature space based on label similarity and density, ensuring minority samples remain distinguishable.  (For general regression, RnC~\cite{zha2023RNC} learns continuous, ranking-aware embeddings by contrasting samples based on their relative ordering in label space. Ordinal Entropy~\cite{zhang2022improving} enforces local ordinal relationships via an entropy-based regularize.)

Loss re-weighting (importance) methods rebalance the importance of each instance during training by adapting the regression objective. Inverse‑frequency weighting is ubiquitous\cite{yang2021ldsfds}; DenseLoss~\cite{steininger2021density} employs label-density weighting calibrated linearly with a tunable parameter $\alpha$. Balanced MSE~\cite{ren2022bmse} derives a closed-form objective assuming uniform label distributions, approximating the resulting integral numerically to address imbalance directly within the regression loss.

For imbalanced (long-tailed) classification, various approaches have been proposed, including resampling  \cite{zhou2020bbn}, logit adjustments \cite{menon2021logitadjustment},  decoupled training \cite{kang2020decoupling}, representation learning \cite{kang2021kcl}, uniform class separation \cite{li2022targeted}, multiple branches \cite{zhou2020bbn}, and multiple experts \cite{zhang2022self}.

\section{Approach}
\label{sec:approach}


{\bf Preliminaries.}  Let $\{(\mathrm{x}_i, y_i)\}_{i=1}^N$ be a training data set $\mathcal{D}$, where $\mathrm{x}_i \in \mathbb{R}^d$ denotes the input of dimensionality $d$ and $y_i \in \mathbb{R}$ is the label, which is a continuous target. We denote $\mathrm{z} = g(\mathrm{x}; \theta_g)$ as the latent representation for $\mathrm{x}$, where $g(\mathrm{x}; \theta_g)$ is the encoder parameterized by a neural network model with parameters $\theta_g$. The final prediction $\hat{y}$ is given by a regressor $f(\mathrm{z}; \theta_f)$ that operates over $\mathrm{z}$, where $f(\cdot; \theta_f)$ is parameterized by $\theta_f$. Therefore, the prediction can be expressed as $\hat{y} = f(g(\mathrm{x}; \theta_g); \theta_f)$.

{\bf Highly imbalanced distributions.}  We divide the targets into equal-width bins.  We define the imbalanced frequency ratio as $\rho = freq_{max}/freq_{min}$, where $freq_{max}$ ($freq_{min})$ is the frequency of instances in the largest (smallest non-empty) bin. A distribution is \emph{highly imbalanced} if $\rho \ge 1000$.

{\bf Estimating probability density distributions.} 
We use Kernel Density Estimation (KDE) \cite{silverman2018density} with a Gaussian kernel. The estimated density $\widehat{p}_Y(y)$ for target value $y$ is:
$
\widehat{p}_Y(y) = \frac{1}{Nh}\sum_j^N K((y - y_j)/h),
$
where $K$ is the kernel and $h$ is the bandwidth.  KDE is used in DenseLoss \cite{steininger2021density} and LDS \cite{yang2021ldsfds}. 
  
We denote $\hat{d_i} \;=\; \widehat{p}_Y(y_i)$. We normalize all densities into $(0,1)$:
$d_i = \hat{d_i} / (\hat{d}_{\max} + \epsilon),$ where $\hat{d}_{\max}$ is the largest density and $\epsilon$ is a small constant ($10^{-3}$) so that $d_i \neq 1$ (and $MDI(d_i) \neq 0$ in Sec.~\ref{sec:mdi}). Henceforth \(d_i\) refers to the normalized density. 
We define the imbalance density ratio as $\rho_d = d_{\max}/d_{\min}$, where $d_{\max} = \max_i d_i$ ($d_{\min} = \min_i d_i,$) is the maximum (minimum) normalized density.  
For KDE, bandwidth $h$ is chosen such that the imbalance density ratio is close to the imbalance frequency ratio; that is, $\rho_d \approx \rho$.

To handle high imbalance, based on KDE, we use the MDI importance function (Sec.~\ref{sec:importance}) to shift importance away from frequent instances and toward rare instances. To consider correlation in addition to error, we use weighted Pearson Correlation Coefficent (wPCC, Sec.~\ref{sec:wpcc}) as a loss regularizer. Stratified Sampling (Sec.~\ref{sec:stratified}) helps produce mini-batches that consistently contain rare samples. Our method is called CISIR, which uses wPCC with MDI in the loss function and stratified sampling for mini-batches (Algorithm~\ref{alg:cisir}).

\subsection{Importance Functions}
\label{sec:importance}
To encourage that both frequent and rare regions of the feature space are learned well, we attribute importance $r_i$ to instance $\mathrm{x}_i$ with normalized density $d_i \in (0,1)$ based on an \emph{importance} ($Imp$) function:
\begin{equation}
    r_i \;=\; Imp(d_i).
\label{eq:importance}
\end{equation}
We use "importance" to distinguish it from a "weight" in the model. The importance function $Imp$ is a monotonically decreasing function: the lower the normalized density $d_i$, the higher the resulting importance $r_i$. The importance can be precomputed once for a dataset and reused during training.

\begin{figure}[t]
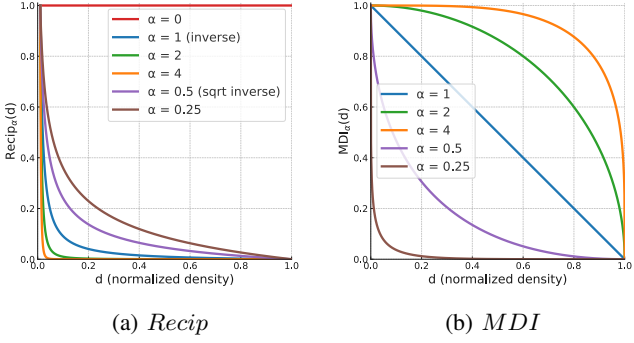

    \centering
    \begin{subfigure}[t]{0.5\columnwidth}
        \centering
        \includesvg[width=\linewidth]{assets/reciprocal_plot.svg}
        \caption{$Recip$}
        \label{fig:recip_infl}
    \end{subfigure}\hfill
    \begin{subfigure}[t]{0.5\columnwidth}
        \centering
        \includesvg[width=\linewidth]{assets/mdi_plot.svg}
        \caption{$MDI$}
        \label{fig:mdi_infl}
    \end{subfigure}
    
    \caption{$Recip$ and $MDI$ importance functions.  For $Recip$, we rescale $r_i=Recip_{\alpha}(d_i)$ so that $r_i\!\in\!(0,1]$ to match $MDI$.}
    \label{fig:importance_fn}
\end{figure}

\subsubsection{Reciprocal Importance}
\label{sec:reciprocal}
The inverse function \cite{yang2021ldsfds} is a typical importance function due to its property of balancing the data distribution.  To reduce the initial sharp decrease in importance at low density, square-root inverse \cite{yang2021ldsfds} was also proposed.  However, the desirable rate of initial decrease is generally not known and dependent on the dataset, imbalance ratio, and loss function.

To overcome these limitations, we generalize inverse and square-root inverse to a \textbf{Reciprocal importance} function that provides more flexibility in the rate of initial decrease in importance at low density. Given the normalized density $d_i$ for instance $\mathrm{x}_i$, we define the reciprocal importance ($Recip$) as:
\begin{equation}
Recip_\alpha(d_i) = \frac{1}{d_i^\alpha}, \quad \alpha \ge 0,
\label{eq:reciprocal}
\end{equation}
where $\alpha$ controls the curvature of the function:
\begin{itemize}
\item When $\alpha = 0$, the function reduces to a constant importance of 1 for all instances; ie, no adjustment for imbalance.
\item When $\alpha = 1$, it is the typical inverse function $1/d_i$ that achieves the balanced distribution.
\item When $\alpha > 1$, it further emphasizes the rare samples beyond the balanced distribution (which is beneficial to some datasets (Sec~\ref{sec:comparing_importance})).
\end{itemize}
Figure \ref{fig:recip_infl} shows Reciprocal importance with various $\alpha$ values.

\subsubsection{Monotonically Decreasing Involution (MDI) Importance}
\label{sec:mdi}
Although Reciprocal importance provides an effective adjustment of importance among frequent and rare instances, it is fundamentally limited by its inherent convexity and exponential form, restricting its ability to represent linear or concave importance relationships. Consequently, Reciprocal importance is insufficient when a more diverse range of importance shapes is needed in various imbalance scenarios.

To address these limitations, we introduce \textbf{Monotonically Decreasing Involution (MDI) importance}, a parameterized function with three properties: (1) it is monotonically decreasing, (2) it allows convex, linear, and concave functions to represent a diverse family of importance functions, and (3) it is an involution (or a self-inverse function, where $f(f(x)) = x$) to preserve the same property exhibited by the inverse function that yields a balanced distribution.
Given a normalized density $d_i$ for instance $\mathrm{x}_i$, MDI importance is defined as:
\begin{equation}
MDI_\alpha(d_i) = \left(1 - d_i^\alpha\right)^{\frac{1}{\alpha}}, \quad \alpha > 0,
\label{eq:mdi}
\end{equation}
where $\alpha$ controls the curvature and shape of the function:
\begin{itemize}
\item When $0 < \alpha < 1$, the function is convex, similar to Reciprocal importance (Sec.~\ref{sec:reciprocal}).
\item When $\alpha = 1$, it is linear, similar to DenseLoss \cite{steininger2021density}.
\item When $\alpha > 1$, it is concave (which is beneficial to some datasets (Sec~\ref{sec:comparing_importance})). 
\item When $\alpha \gg 1$, it is approximately 1 at low density. 
\end{itemize}

Fig.~\ref{fig:mdi_infl} illustrates MDI importance with various $\alpha$ values. Like learning rate, $\alpha$ is selected based on the validation set.

\subsection{Weighted Pearson Correlation Coefficient (wPCC) as Loss Regularization}
\label{sec:wpcc}
For regression, as discussed in Sec.~\ref{sec:intro}, MSE does not differentiate models that are equally inaccurate but differ in  correlation between the predicted and actual values.  Also, while perfect MSE yields perfect positive correlation, perfect positive correlation does not guarantee perfect MSE.  
Hence, for imbalanced regression, we propose $wMSE$ (weighted MSE) as the primary loss function and $wPCC$ (weighted Pearson Correlation Coefficient) as the secondary loss function or regularization.  Each instance $i$ is weighted by importance $r_i$ (Eq.~\ref{eq:importance}), which has been normalized such that $\sum_{i=1}^{N} r_i = 1$. To allow different importance values (e.g. different $\alpha$ values in $MDI$ for $wMSE$ and $wPCC$) for the same instance in the two loss functions, we denote $re_i$ and $rc_i$ as the importance for instance $i$ in $wMSE$ and $wPCC$ respectively.  Moreover, $re_i$ and $rc_i$ are obtained from an importance function (Recip or MDI in Sec.~\ref{sec:importance}) with $\alpha$ values that we denote as $\alpha_e$ and $\alpha_c$ respectively.

We define $wMSE$ as:
\begin{equation}
    wMSE
    \;=\;
    \sum_{i=1}^{N} re_i\,(y_i-\hat{y}_i)^{2},
    \label{eq:wmse}
\end{equation}
and $wPCC$ as:
\begin{equation}
wPCC
\;=\;
1 - \frac{
       \displaystyle\sum_{i=1}^{N}
       rc_i\,(y_i-\bar{y})\,(\hat{y}_i-\bar{\hat{y}})
     }
     {%
       \sqrt{\displaystyle\sum_{i=1}^{N} rc_i\,(y_i-\bar{y})^{2}}\;
       \sqrt{\displaystyle\sum_{i=1}^{N} rc_i\,(\hat{y}_i-\bar{\hat{y}})^{2}}
     },
\label{eq:wpcc}
\end{equation}
where $\bar{y}$ and $\bar{\hat{y}}$ are the averages of $y$ and $\hat{y}$ respectively.
Our proposed overall loss function is:
\begin{equation}
\mathcal{L} 
\;=\;
wMSE \;+\; \lambda \cdot wPCC,
\label{eq:loss}
\end{equation}
where $\lambda>0$ adjusts the influence of $wPCC$.

Moreover, $wPCC$ can help reduce $wMSE$. Following the MSE decomposition in \cite[Eq.~9]{murphy1988skill}, we have:
\begin{align}
\text{MSE}(\hat{y},y)
  &= \bigl(\bar{y} - \bar{\hat{y}}\bigr)^{2}
   + \operatorname{var}(\hat{y})
   + \operatorname{var}(y) \notag \\
  &\quad - 2\cdot\operatorname{cov}(\hat{y}, y),
\label{eq:bv-unw}
\end{align}
which, after some transformation, as shown in Appendix~\ref{app:mse-decomp}, yields:
\begin{align}
\text{MSE}(\hat{y},y)
  &= \bigl(\bar{y} - \bar{\hat{y}}\bigr)^{2}
   + \bigl(\operatorname{sd}(\hat{y}) - \operatorname{sd}(y)\bigr)^{2} \notag \\
  &\quad + 2\,\operatorname{sd}(\hat{y})\,\operatorname{sd}(y)\,
     \Bigl(1 - \operatorname{PCC}(\hat{y},y)\Bigr).
\label{eq:bv-sd}
\end{align}
where $\operatorname{var}(\cdot)$, $\operatorname{cov}(\cdot,\cdot)$, and $\operatorname{sd}(\cdot)$ denote the variance, covariance, and standard deviation operators, respectively. 

Eq.~\eqref{eq:bv-sd} is a sum of three terms: a term for mismatch in the mean, $(\bar{y} - \bar{\hat{y}})^{2}$; a second term for mismatch in the standard deviation, $(\operatorname{sd}(\hat{y}) - \operatorname{sd}(y))^{2}$; and a third term, $2\,\operatorname{sd}(\hat{y})\,\operatorname{sd}(y)\,(1 - \operatorname{PCC}(\hat{y},y))$, related to the correlation deficit. While the first two terms encourage matching the moments of the data distribution, the third term's minimization highlights a critical point. MSE can be reduced by increasing the Pearson Correlation Coefficient, $\operatorname{PCC}(\hat{y},y)$, towards $1$, and/or by decreasing the standard deviation of the predictions, $\operatorname{sd}(\hat{y})$ towards 0.

In general, since the second term is present and $\operatorname{sd}(y) \neq 0$, the prediction standard deviation $\operatorname{sd}(\hat{y})$ does not collapse to zero. 
However, in some cases, illustrated by Model 1 (with no correlation) and
Model 2 (with some correlation) in Fig.~\ref{fig:threemodels}, MSE cannot distinguish the
two models.  This is caused by $\operatorname{sd}(\hat{y})$ of Model 1 being zero, which
renders the third term to be zero and the lack of correlation to be
ignored.  Similarly, a small $\operatorname{sd}(\hat{y})$ diminishes the penalty due to
poor correlation.
To address this limitation, our proposed $wPCC$ regularizer directly penalizes poor correlation by minimizing $1 - \operatorname{PCC}(\hat{y},y)$. 

\subsection{Stratified Sampling in Mini-Batches (SSB)}
\label{sec:stratified}

Training neural networks with stochastic gradient descent (SGD) typically involves partitioning the dataset \(\mathcal{D}\) into mini-batches of size \(B\), yielding \(M = N/B\) mini-batches and parameter updates per epoch. With highly imbalanced data, uniformly sampled mini-batches may not represent rare target regions adequately.  Consider the probability of rare samples as $\pi_r \approx 0$, the probability of no rare instances in a uniformly drawn mini-batch is $(1-\pi_r)^B$, which is close to 1 when \(B\) is relatively small compared to \(1/\pi_r\).  That is, some mini-batches might not have rare instances at all, which lead to gradients that do not reduce loss for rare instances during some model updates.

To ensure rare instances are represented across mini-batches, we propose to perform stratified sampling to form mini-batches such that each mini-batch has
a similar distribution as the overall training distribution.  Consequently, gradients computed from these stratified mini-batches approximate the gradient computed over the training set, and every model update reduces loss for some rare instances.

To perform stratified sampling, we  first choose $M$ such that $M$ is less than or equal to the number of rare instances.  We sort all instances based on their target values.  The sorted instances are then divided into $B$ groups, each containing $M$ instances (The final group may have fewer than $M$ instances). The number of groups $B$ matches the size of a mini-batch, with each group contributing one instance to the mini-batch. To create each mini-batch, we randomly select one instance from each of the $B$ groups.  For example, if larger target values are rarer, the $M$ instances with the largest target values are in one group.  Each of the $M$ instances is randomly assigned to one of $M$ mini-batches.

\begin{algorithm}[t]
\caption{CISIR Training Procedure}
\label{alg:cisir}
\begin{algorithmic}[1]
\Require Dataset $\mathcal{D}$, batch size $B$, hyperparameters
         $\alpha_e,\alpha_c,\lambda$, learning-rate $\eta$
\State Estimate densities $d_i$ with KDE (Sec.~\ref{sec:approach})
\State Compute importance $re_i,rc_i$ using MDI (Eq.~\eqref{eq:mdi})
\State Form $B$ groups with Stratified Sampling (Sec.~\ref{sec:stratified})
\While{model not converged}
    \State Build mini-batch $\mathcal{B}$ by drawing one sample
           from each group
    \State Forward-propagate to obtain predictions $\hat{y}_i$
    \State Compute $wMSE$ (Eq.~\eqref{eq:wmse}) and $wPCC$ (Eq.~\eqref{eq:wpcc})
    \State $\mathcal{L} \gets wMSE + \lambda \cdot wPCC$ \hfill (Eq.~\eqref{eq:loss})
    \State Back-propagate $\nabla_\theta \mathcal{L}$ and update model parameters
          with step $\eta$
\EndWhile \\
\Return Trained model parameters $\theta$
\end{algorithmic}
\end{algorithm}

Algorithm \ref{alg:cisir} summarizes the overall CISIR method. We estimate densities via KDE on line 1. We then compute the importance values using MDI on line 2. With stratified sampling, we form B groups of data instances on line 3. From lines 4 to 9, we iteratively train the model by getting a mini-batch on line 5, taking the model's predictions on line 6, computing the loss with lines 7-8, and updating the model parameters on line 9. After convergence, the trained model is finally returned on line 11.

\renewcommand{\arraystretch}{1.25}
\begin{table*}[t]
  \caption{CISIR vs.\ recent methods ({\bfseries bold} = best, \underline{underline} = 2nd best, * = statistically significant)}
  \label{tab:main-results}
  \centering
  \small                                
  \setlength{\tabcolsep}{3pt}           
  \renewcommand{\arraystretch}{1.10}    
  \begin{tabular*}{0.8\textwidth}{@{\extracolsep{\fill}} l l c c | c | c c | c |}
    \toprule
    Dataset & Method & MAE$\downarrow$ & MAE$_{\mathrm{R}}\downarrow$ & \multicolumn{1}{c|}{AORE$\downarrow$} & PCC$\uparrow$ & PCC$_{\mathrm{R}}\uparrow$ & \multicolumn{1}{c|}{AORC$\uparrow$} \\
    \midrule
    \multirow{5}{*}{\textbf{SEP-EC}}
      & SQINV+LDS+FDS  & 0.177 & \underline{0.566}* & 0.371 & -0.025 & 0.067 & 0.021 \\
      & BalancedMSE    & 0.161 & 0.659 & 0.410 & -0.036 & 0.141 & 0.053 \\
      & DenseLoss      & \textbf{0.071}* & 0.626 & \underline{0.348}* & \textbf{0.286} & \underline{0.699}* & \textbf{0.493} \\
      & Recip+wPCC+SSB & \underline{0.089}* & 0.606 & \underline{0.348}* & 0.219 & \underline{0.699}* & 0.459 \\
      & CISIR          & 0.184 & \textbf{0.441}* & \textbf{0.313}* & \underline{0.274}* & \textbf{0.703} & \underline{0.488}* \\
    \midrule
    \multirow{5}{*}{\textbf{SEP-C}}
      & SQINV+LDS+FDS  & 1.681 & 4.314 & 2.997 & 0.173 & 0.481 & 0.327 \\
      & BalancedMSE    & 1.683 & 3.614 & 2.649 & 0.394 & 0.393 & 0.393 \\
      & DenseLoss      & \textbf{0.237}* & 2.245 & \underline{1.241} & \underline{0.690}* & 0.588 & 0.639 \\
      & Recip+wPCC+SSB & 1.173 & \textbf{1.376}* & 1.274 & 0.627 & \textbf{0.661}* & \underline{0.644} \\
      & CISIR          & \underline{0.335}* & \underline{1.875}* & \textbf{1.105}* & \textbf{0.702} & \underline{0.593} & \textbf{0.647} \\
    \midrule
    \multirow{5}{*}{\textbf{SARCOS}}
      & SQINV+LDS+FDS  & 0.575 & 0.748 & 0.661 & 0.020 & -0.049 & -0.015 \\
      & BalancedMSE    & 0.571 & 1.694 & 1.132 & 0.189 & -0.170 & 0.010 \\
      & DenseLoss      & 0.058 & 0.076 & 0.067 & 0.964 & 0.830 & 0.897 \\
      & Recip+wPCC+SSB & \textbf{0.053} & \underline{0.071} & \textbf{0.062} & \textbf{0.986} & \textbf{0.910}* & \textbf{0.948}* \\
      & CISIR          & \underline{0.055} & \textbf{0.069} & \textbf{0.062} & \underline{0.982}* & \underline{0.876}* & \underline{0.929}* \\
    \midrule
    \multirow{5}{*}{\textbf{BF}}
      & SQINV+LDS+FDS  & 1.036 & 1.780 & 1.408 & -0.152 & 0.065 & -0.044 \\
      & BalancedMSE    & 0.689 & 1.769 & 1.229 & -0.011 & 0.066 & 0.028 \\
      & DenseLoss      & \textbf{0.169}* & 0.747 & \textbf{0.458} & \underline{0.735} & 0.301 & 0.518 \\
      & Recip+wPCC+SSB & \underline{0.187}* & \underline{0.740} & \underline{0.463}* & 0.733 & \underline{0.319}* & \underline{0.526} \\
      & CISIR          & 0.280 & \textbf{0.709}* & 0.495 & \textbf{0.737} & \textbf{0.330}* & \textbf{0.533} \\
    \midrule
    \multirow{5}{*}{\textbf{ONP}}
      & SQINV+LDS+FDS  & 2.628 & 4.438 & 3.533 & 0.034 & -0.012 & 0.011 \\
      & BalancedMSE    & 2.798 & 4.154 & 3.476 & -0.028 & 0.012 & 0.047 \\
      & DenseLoss      & \textbf{0.317} & \underline{1.311} & \underline{0.814} & \textbf{0.325}* & 0.054 & 0.189 \\
      & Recip+wPCC+SSB & \underline{0.326} & 1.351 & 0.838 & 0.288 & \textbf{0.095} & \underline{0.192} \\
      & CISIR          & 0.379 & \textbf{1.180}* & \textbf{0.780} & \underline{0.299}* & \underline{0.093}* & \textbf{0.196} \\
    \bottomrule
  \end{tabular*}
\end{table*}

\begin{table*}[t]
  \caption{Incorporating wPCC into other methods ({\bf bold} = best, * = statistically significant)}
  \label{tab:wPCC_results}
  \centering
  \small                
  \setlength{\tabcolsep}{3pt}
  \begin{tabular}{l*{5}{>{\centering\arraybackslash}p{1.15cm} >
                      {\centering\arraybackslash}p{1.15cm}}}
    \toprule
    \multirow{2}{*}{Method} &
    \multicolumn{2}{c}{SEP-EC} &
    \multicolumn{2}{c}{SEP-C}  &
    \multicolumn{2}{c}{SARCOS} &
    \multicolumn{2}{c}{BF}     &
    \multicolumn{2}{c}{ONP} \\
    \cmidrule(lr){2-3}\cmidrule(lr){4-5}\cmidrule(lr){6-7}\cmidrule(lr){8-9}\cmidrule(lr){10-11}
      & AORE$\!\downarrow$ & AORC$\!\uparrow$
      & AORE$\!\downarrow$ & AORC$\!\uparrow$
      & AORE$\!\downarrow$ & AORC$\!\uparrow$
      & AORE$\!\downarrow$ & AORC$\!\uparrow$
      & AORE$\!\downarrow$ & AORC$\!\uparrow$ \\
    \midrule
    SQINV+LDS+FDS  & \textbf{0.371} & 0.021 & 2.997 & 0.327 & 0.661 & $-0.015$ & 1.408 & $-0.044$ & 3.533 & 0.011 \\
    \;+\;wPCC      & \textbf{0.371} & \textbf{0.270*} & \textbf{2.891} & \textbf{0.411*} & \textbf{0.657} & \textbf{0.051*} & \textbf{1.062*} & \textbf{0.130*} & \textbf{3.416*} & \textbf{0.038} \\
    \specialrule{.01pt}{1pt}{1pt}
    BalancedMSE    & 0.410 & 0.053 & 2.649 & 0.393 & 1.132 & 0.010 & 1.229 & 0.028 & 3.476 & 0.047 \\
    \;+\;wPCC      & \textbf{0.392} & \textbf{0.057} & \textbf{2.484} & \textbf{0.491*} & \textbf{0.884*} & \textbf{0.097*} & \textbf{0.990} & \textbf{0.162*} & \textbf{3.223*} & \textbf{0.086*} \\
    \specialrule{.01pt}{1pt}{1pt}
    DenseLoss      & 0.348 & 0.493 & 1.241 & 0.639 & 0.067 & 0.897 & \textbf{0.458} & 0.518 & 0.814 & 0.189 \\
    \;+\;wPCC      & \textbf{0.344} & \textbf{0.500} & \textbf{1.084} & \textbf{0.660} & \textbf{0.062} & \textbf{0.920*} & \textbf{0.458} & \textbf{0.529} & \textbf{0.791} & \textbf{0.203} \\
    \bottomrule
  \end{tabular}
\end{table*}

\section{Experimental Evaluation}
\label{sec:eval}
\subsection{Datasets and Evaluation Metrics}
\label{sec:dataset_metrics}
We evaluate our method on five highly imbalanced datasets. SEP-EC aims to forecast the change (delta) in proton intensity based on features from electron intensity and CMEs (Coronal Mass Ejections). SEP-C focuses on forecasting peak proton intensity based on CMEs characteristics. SARCOS \cite{vijayakumar2000sarcos} estimates the torque vector based on joint-state inputs for a 7‑DOF robot arm. Blog Feedback (BF) \cite{buza2014bf} forecasts the number of comments based on textual, temporal, and engagement features.
Online News Popularity (ONP) \cite{fernandes2015onp} estimates the number of shares of an article based on content, topic, and sentiment attributes.  Each dataset's imbalance ratio ($\rho$) is in Table~\ref{tab:importance-results}. SARCOS, BF, and ONP datasets are available at \cite{vijayakumar2000sarcos_data, buza2014blogfeedback_data, fernandes2015onp_data}. Our SEP datasets are available in \href{https://huggingface.co/datasets/Machine-Earning/CISIR-datasets/resolve/main/CISIR-data.zip}{https://huggingface.co/datasets/Machine-Earning/CISIR-datasets/resolve/main/CISIR-data.zip}.

Our evaluation metrics on the test set include overall $MAE$ (Mean Absolute Error) and $PCC$ (Pearson Correlation Coefficient).  $MAE_R$ ($PCC_R$) denotes $MAE$ ($PCC$) of rare instances that are important.  For SEP-EC, only positive (increasing intensity) rare instances are included in $MAE_R$ and $PCC_R$.  Since we would like to consider all instances with an emphasis on rare instances, we propose two hybrid metrics.  We define the average of overall and rare $MAE$ as: $AORE = (MAE + MAE_R)/2$, and overall and rare $PCC$ as: $AORC = (PCC + PCC_R)/2$.  $AORE$ and $AORC$ are our main evaluation metrics, with $AORE$ being more important than $AORC$ (as we discussed in Sec.\ref{sec:intro} that error is more important than correlation).
To determine if two averages of measurements are different with statistical significance, we use standard error.  

\subsection{Baseline Methods and Experimental Procedures}
\label{sec:exp_proc}

Since our proposed CISIR is most similar to DenseLoss \cite{steininger2021density}, LDS/FDS \cite{yang2021ldsfds}, and Balanced MSE \cite{ren2022bmse}, they are the baseline methods.  We do not include methods that need substantive additional computation due to factors such as additional synthetic instances and separate representation learning.  
For LDS/FDS, we choose SQINV+LDS+FDS, which is the more effective variant \cite{yang2021ldsfds}.  Also, we include a variant of CISIR denoted as Recip+wPCC+SSB, which uses Recip instead of MDI.  The baseline results are obtained by implementing DenseLoss ourselves, while using the authors' implementations, including their hyperparameters, of LDS/FDS~\cite{yang2021ldsfds_github} and Balanced-MSE~\cite{ren2022bmse_github}. Our source code is available in \href{https://github.com/Machine-Earning/CISIR}{https://github.com/Machine-Earning/CISIR}.

All experiments are implemented in TensorFlow 2.6 and executed on NVIDIA RTX 2080Ti (11GB), A100 (40GB), and H100 (80GB) GPUs. Each experiment is repeated with five fixed seeds;
reported results are averages over these runs. For all datasets, we use a residual MLP architecture with LeakyReLU activations, batch normalization, with dataset-specific hidden units, embedding dimensions, dropout rates, weight-decay values, batch sizes, and initial learning rates. Models are trained till convergence with Adam optimizer, early stopping based on validation loss, and a learning-rate scheduler reducing by a factor of 0.95 after 50 stagnant epochs. If training and testing subsets are provided, we use them directly; otherwise, we adopt a 2/3–1/3 split for training and testing with stratified sampling, performing four-fold cross-validation on the training portion, again with stratified sampling.

\renewcommand{\arraystretch}{1.25}
\begin{table}[t]
  \caption{Comparison of importance functions ({\bf bold} = best, \underline{underline} = 2nd best, * = statistically significant)}
  \label{tab:importance-results}
  \centering
  \begin{tabular}{lccccc}
    \toprule
    \multirow{2}{*}{\makecell{Importance\\functions}} & \multicolumn{5}{c}{AORE↓} \\
    \cmidrule(l){2-6}
    & SEP-C & SARCOS & ONP & SEP-EC & BF \\

    \cmidrule(l){2-6}
     Imb. ratio $\rho$ & \textit{1{,}476} & \textit{3{,}267} & \textit{6{,}746} & \textit{10{,}478} & \textit{33{,}559} \\
    \midrule
    INV        & 2.456 & 0.064 & 0.849 & 0.362 & 0.451 \\
    SQINV      & 1.313 & 0.068 & \underline{0.765} & 0.369 & \underline{0.447} \\
    DenseLoss  & \underline{1.277} & 0.065 & 0.842 & 0.355 & 0.471 \\
    Recip      & 1.289 & \underline{0.063} & \textbf{0.764} & \underline{0.349} & \textbf{0.445} \\
    MDI        & \textbf{1.105}* & \textbf{0.062} & 0.780 & \textbf{0.313}* & 0.495 \\ 
    \hline\hline 
    $\alpha_e$ in Recip & 0.70 & 1.10 & 0.90 & 1.20 & 0.70\\
    $\alpha_e$ in MDI   & 2.40 & 0.20 & 1.10 & 0.01 & 0.50\\
    \bottomrule
  \end{tabular}
\end{table}

\begin{figure*}[!t]      
  \centering
  \includegraphics[width=\textwidth]{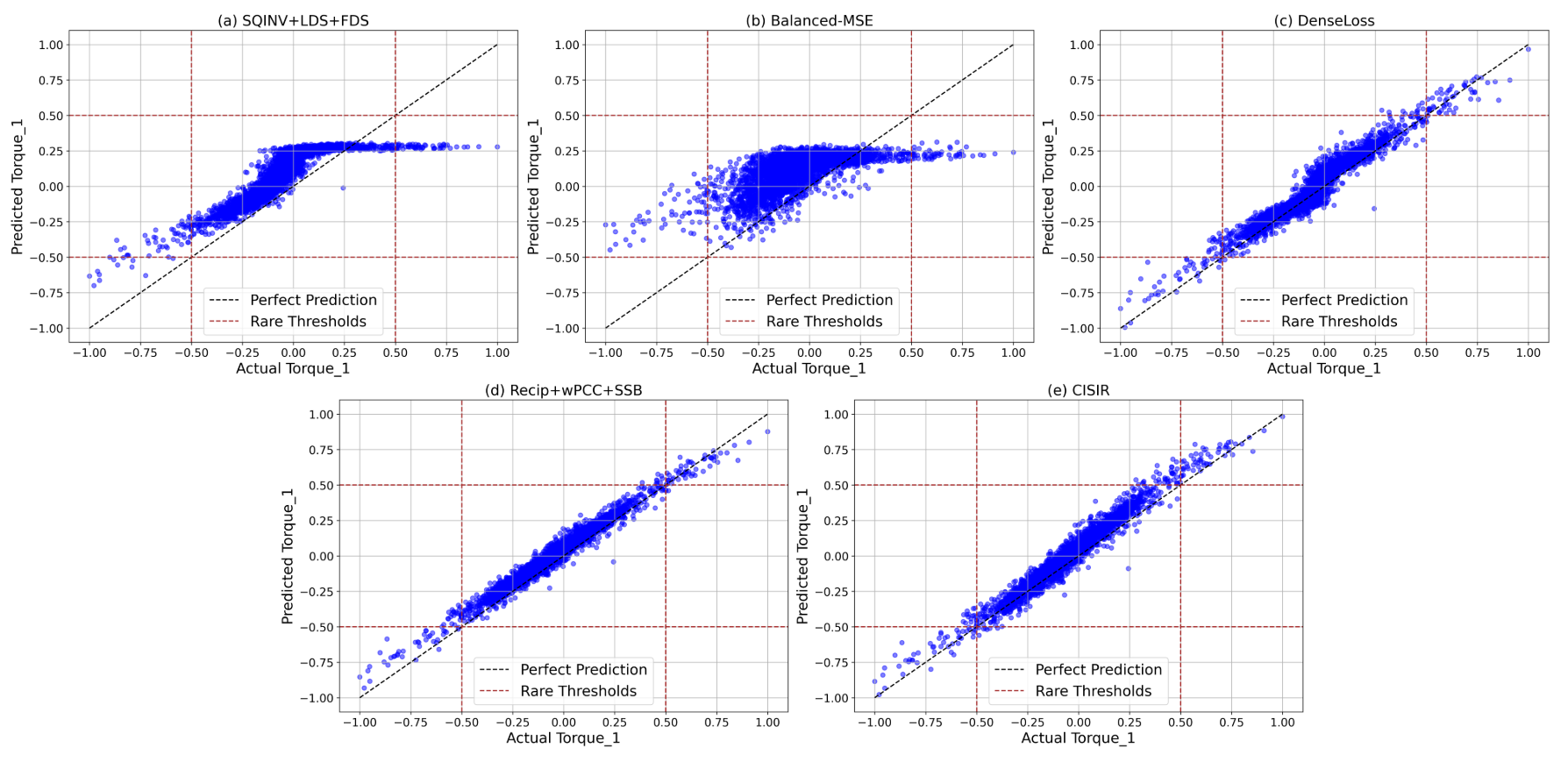}
  \caption{Actual vs.\ predicted values on the SARCOS dataset.}
  \label{fig:avsp_five}
\end{figure*}

\renewcommand{\arraystretch}{1.15} 
\begin{table}[t]
  \caption{Ablation study of CISIR on the SEP-EC dataset ({\bf bold} = best, \underline{underline} = 2nd best)}
  \label{tab:ablation}
  \centering
  \begin{adjustbox}{center,max width=\linewidth}
  \setlength{\tabcolsep}{6pt}
  \begin{tabular}{lccrr}
    \toprule
    \multirow{2}{*}{Method} &
    \multicolumn{1}{c}{AORE$\downarrow$} &
    \multicolumn{1}{c}{AORC$\uparrow$} \\[2pt]
    \midrule
    \rowcolor{gray!10}
    CISIR & 0.313 & 0.488 \\  
    w/o MDI (with INV)          & 0.362 (\underline{+0.049}) & 0.447 (\underline{-0.041}) \\
    w/o wPCC                & 1.160 (\textbf{+0.847}) & 0.194 (\textbf{-0.294}) \\
     w/o SSB                & 0.333 (+0.020)  & 0.463 (-0.025) \\
    \bottomrule
  \end{tabular}
  \end{adjustbox}
\end{table}

\subsection{Results}
Table~\ref{tab:main-results} has our main results in comparing CISIR to other recent methods.  We observe that CISIR achieves the lowest AORE in 4 datasets, while it achieves the highest AORC in 3 and the second highest in the remaining 2 datasets.  That is, generally, CISIR outperforms the other methods based on our main evaluation metrics.
Focusing on important rare instances, CISIR has the lowest $MAE_R$ in 4 and the second lowest in the remaining dataset.  While CISIR has the highest $PCC_R$ in 2 datasets, Recip+wPCC+SSB (a variant of CISIR) has the highest in 3 datasets.  That is, CISIR and its variant generally outperform the other methods on rare instances.  

{\bf Incorporating wPCC into other methods.}
To evaluate how wPCC can benefit other methods, we incorporate it into other methods.  According to Table~\ref{tab:wPCC_results}, wPCC mostly improves all 3 methods across all 5 datasets in both AORE and AORC. The only exceptions are SQINV+LDS+FDS on SEP-EC and DenseLoss on BF, where AORC improved while AORE did not change.  That is, generally, wPCC can improve the performance of other methods.  This indicates that the constraint from wPCC in the loss not only improves correlation, but also reduces error.

{\bf Comparing importance functions.}
\label{sec:comparing_importance}
To compare importance functions directly, we use and fix wPCC (Eq.~\ref{eq:wpcc}) and SSB (Sec.~\ref{sec:stratified}), but only vary importance functions for wMSE (Eq.~\ref{eq:wmse}).  INV is the inverse and SQINV is the square-root inverse \cite{yang2021ldsfds}. Table~\ref{tab:importance-results} shows the AORE results. Since we fix wPCC, AORC is not directly affected and is not included in the table.  MDI achieves the lowest AORE in 3 datasets and Recip in 2 datasets.  That is, MDI generally outperforms the other importance functions.

To analyze if the imbalance ratio ($\rho$) of a dataset is related to  $\alpha$ in Recip (Eq.~\ref{eq:reciprocal}) or MDI (Eq.~\ref{eq:mdi}), we include $\rho$ and $\alpha_e$ (which is used to obtain importance $re_i$ for wMSE in Eq.~\ref{eq:wmse}) in Table~\ref{tab:importance-results}.  However, we do not observe a meaningful relationship between $\rho$ and $\alpha_e$.
Unlike the other importance functions, we design MDI to include concave functions when $\alpha > 1$ (Eq.~\ref{eq:mdi}).  From Table~\ref{tab:importance-results}, we observe that the SEP-C and ONP datasets benefit from concave functions with $\alpha_e > 1$.  Also, we design Recip to allow further emphasis on rare instances beyond the balanced distribution when $\alpha > 1$ (Eq.~\ref{eq:reciprocal}).  We observe that the SARCOS and SEP-EC datasets benefit from further emphasis on rare instances with $\alpha_e > 1$.

Based on the observations above, the weaker performance of SQINV+LDS+FDS and Balanced MSE in Table~\ref{tab:main-results} may be due to the absence of wPCC, which can reduce error and increase correlation.  Also, SQINV is equivalent to Recip with $\alpha=0.5$, but $\alpha_e$ for Recip in Table~\ref{tab:importance-results} can be quite different.  Our datasets are highly imbalanced, but their datasets might not be (imbalance ratios were not usually reported). While LDS uses KDE, it does not explicitly maintain the imbalance ratio ($\rho_d \approx \rho$) in KDE (Sec.\ref{sec:approach}), which might be important for high imbalance ratios.

\begin{figure}[t]
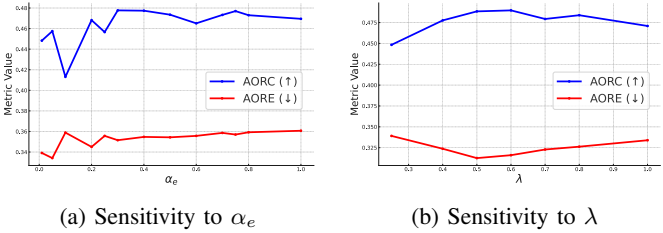

  \centering
  \begin{subfigure}[t]{0.48\linewidth}
    \centering
    \includesvg[width=\linewidth]{assets/aore_aorc_vs_alphae.svg}
    \caption{Sensitivity to $\alpha_e$}
    \label{fig:alphae_sensitivity}
  \end{subfigure}\hfill
  \begin{subfigure}[t]{0.48\linewidth}
    \centering
    \includesvg[width=\linewidth]{assets/aore_aorc_vs_lambda.svg}
    \caption{Sensitivity to $\lambda$}
    \label{fig:lambda_sensitivity}
  \end{subfigure}
  \caption{Sensitivity analysis of CISIR to $\alpha_e$ and $\lambda$ on SEP-EC.}
  \label{fig:sensitivity_cisir}
\end{figure}

\subsection{Analyses}

\begin{table*}[t]
\caption{Effects of $w\mathrm{PCC}$ on CISIR's MSE decomposition terms for SEP--EC. CISIR-wPCC denotes CISIR without $w\mathrm{PCC}$.}
\label{tab:mse-decomp-sep-ec}
\centering
\small
\setlength{\tabcolsep}{3pt}
\begin{tabular}{lccccccccc|ccc}
\toprule
\multirow{2}{*}{\shortstack{Test\\Subset}} &
\multicolumn{3}{c}{$\bigl(\bar{y}-\bar{\hat{y}}\bigr)^2$} &
\multicolumn{3}{c}{$\bigl(\operatorname{sd}(\hat{y})-\operatorname{sd}(y)\bigr)^2$} &
\multicolumn{3}{c|}{$2\,\operatorname{sd}(\hat{y})\,\operatorname{sd}(y)\,\bigl(1-\operatorname{PCC}(\hat{y},y)\bigr)$} &
\multicolumn{3}{c}{$1-\operatorname{PCC}(\hat{y},y)$} \\
\cmidrule(lr){2-4}\cmidrule(lr){5-7}\cmidrule(lr){8-10}\cmidrule(lr){11-13}
& CISIR-wPCC & CISIR & \textbf{\%$\Delta$} &
  CISIR-wPCC & CISIR & \textbf{\%$\Delta$} &
  CISIR-wPCC & CISIR & \textbf{\%$\Delta$} &
  CISIR-wPCC & CISIR & \textbf{\%$\Delta$} \\
\midrule
All  & 0.145 & 0.024 & \textbf{-83.3} & 
       1.199 & 0.000 & \textbf{-100.0} & 
       0.304 & 0.025 & \textbf{-91.6} &
       0.963 & 0.726 & \textbf{-24.5} \\
Rare & 0.366 & 0.149 & \textbf{-59.2} & 
       0.786 & 0.027 & \textbf{-96.5} & 
       0.367 & 0.061 & \textbf{-83.3} &
       0.649 & 0.297 & \textbf{-54.3} \\
\bottomrule
\end{tabular}
\end{table*}

To further analyze the main results on the SARCOS dataset, we plot the actual vs. predicted values in Figure~\ref{fig:avsp_five}.  
Rare instances are in the top right and bottom left corners. Both SQINV+LDS+FDS and Balanced MSE have more difficulty with rare instances than the other 3 methods, which perform similarly.  Focusing on the the top right corner, Recip+WPCC+SSB is more accurate.  However, focusing on the few rarest instances on the far top right corner, CISIR is more accurate.  Focusing on the bottom left corner, CISIR is more accurate.

{\bf Ablation study.}
Results from an ablation study on CISIR with the SEP-EC dataset are in Table~\ref{tab:ablation}.  We observe that each of the 3 proposed components (MDI, wPCC, and SSB) contributes to CISIR.  Particularly, wPCC contributes the most. Moreover, wPCC not only increases correlation (AORC), it also reduces error (AORE). This indicates that the constraint from wPCC in the loss helps achieve a lower local minimum for wMSE during training.

{\bf Three components of MSE decomposition.}

Following Eq.~\eqref{eq:bv-sd}, MSE decomposes into three terms: the first term $(\bar{y} - \bar{\hat{y}})^2$ captures mean mismatch, the second term $(\operatorname{sd}(\hat{y}) - \operatorname{sd}(y))^2$ captures standard deviation mismatch, and the third term $2\,\operatorname{sd}(\hat{y})\,\operatorname{sd}(y)\,(1-\operatorname{PCC}(\hat{y},y))$ captures correlation deficit. We analyze how including $wPCC$ regularization affects these three components on the SEP-EC dataset. Since $wPCC$ directly optimizes correlation by minimizing $1-\operatorname{PCC}(\hat{y},y)$, we expect it to primarily reduce the third term. Table~\ref{tab:mse-decomp-sep-ec} shows the decomposition results for CISIR with and without $wPCC$ on both the full test set (\emph{All}) and rare events (\emph{Rare}). Also, $1-\operatorname{PCC}$ is shown separately in the last column.  As expected, $1-\operatorname{PCC}$ and the third term are reduced. However, the first two terms are also reduced.

Without $wPCC$, the second term is the largest contributor to total MSE (1.199 for \emph{All}, 0.786 for \emph{Rare}), dominating both the first term (0.145 and 0.366) and third term (0.304 and 0.367). Including $wPCC$ produces reductions across all three components: $-83.3\%$ (first term), $-100.0\%$ (second term), and $-91.6\%$ (third term) for the full test set, and similarly for rare events ($-59.2\%$, $-96.5\%$, and $-83.3\%$ respectively). The near-complete elimination of the second term is particularly notable, reducing from 1.199 to 0.000 for all samples and from 0.786 to 0.027 for rare events. This suggests that $wPCC$ regularization constrains the model to produce predictions with variability matching the true distribution, addressing the issues inherent in MSE-only optimization.

The third term reductions ($-91.6\%$ for \emph{All}, $-83.3\%$ for \emph{Rare}), directly reflect the intended effect of $wPCC$ on correlation. To isolate the pure correlation improvement from the standard deviation weighting in the third term, we additionally report $1-\operatorname{PCC}(\hat{y},y)$ as a correlation degradation metric. The $wPCC$ regularizer reduces correlation degradation by $-24.5\%$ for all samples (from 0.963 to 0.726, corresponding to PCC improving from 0.037 to 0.274) and by $-54.3\%$ for rare events (from 0.649 to 0.297, corresponding to PCC improving from 0.351 to 0.703). The stronger improvement for rare events demonstrates that $wPCC$ is effective at maintaining correlation where it matters most.

The first term, while showing the smallest absolute values, still improves by $-83.3\%$ for the full test set, and by $-59.2\%$ for the rare event subset. Overall, the decomposition reveals that $wPCC$ addresses multiple aspects of prediction quality simultaneously, with its most pronounced effect being the correction of prediction standard deviation mismatch (second term), followed by improved correlation (third term and $1-\operatorname{PCC}$) and reduced bias (first term).

{\bf Parameter Sensitivity.}
In CISIR, $\alpha_e$ ($\alpha_c$) is used to vary the MDI importance function (Eq.~\ref{eq:mdi}), which in turn varies the importance to instances in wMSE (wPCC).  $\lambda$ varies the contribution of wPCC to the overall loss (Eq.\ref{eq:loss}). 

From Figure~\ref{fig:alphae_sensitivity}, we observe that AORE generally degrades, while AORC generally improves, when $\alpha_e$ increases. That is, AORE and AORC are trading off when $\alpha_e$ varies.
Since $\alpha_e$ affects wMSE (which affects AORE), we prefer $\alpha_e < 0.1$ (which indicates large importance from rare instances).

From Sec.~\ref{sec:wpcc}, $\alpha_c$ governs the importance $rc_i$. Empirically, $rc_i = 1$ (\emph{i.e.}, not using MDI and assigning equal importance to every instance) already enables wPCC, together with an appropriately chosen $\lambda$, to lower error and raise correlation in almost all scenarios. Across the five datasets, only SEP-C needed $\alpha_c$ to be set to 1.7 (which is another example where an "atypical" concave function is beneficial).

From Figure~\ref{fig:lambda_sensitivity}, we observe that AORE bottoms and AORC peaks when $\lambda$ is in the range of 0.5 -- 0.6.    We note that AORE and AORC generally improve and degrade together. That is, AORE and AORC are not trading off and can be optimized together by varying $\lambda$. 
Also, this provides further evidence that wPCC (controlled by $\lambda$) is beneficial in  not only increasing correlation, but also reducing error.

\section{Conclusion, Limitations, and Broader Impacts}
\label{sec:conclusion}
For highly imbalanced regression with tabular data, we propose CISIR that incorporates wPCC as a secondary loss function, MDI importance that allows convex, linear, and concave functions, and stratified sampling in the mini-batches.  Our experimental results indicate that CISIR can achieve lower error and higher correlation than some recent methods.  Also, adding our wPCC component to other methods is beneficial in not only improving correlation, but also reducing error.  Lastly, MDI importance can outperform other importance functions.

While MDI importance allows a linear function, it does not allow the slope to vary (which involves another parameter) as DenseLoss \cite{steininger2021density} does.  The claims of this study are limited to the included methods and datasets.  While our proposed wPCC component is for imbalanced regression, our rationale in Sec.~\ref{sec:intro} is for general regression, so wPCC can benefit regression in general.
Our proposed CISIR can benefit other applications; e.g., estimating rare large credit card fraudulent transactions and rare severe outcomes of diseases.  We do not anticipate negative societal impacts.

\bibliographystyle{plain} 
\bibliography{references}

\appendix

\section{Technical Appendices and Supplementary Material}

\subsection{Involution in MDI influence}
\label{app:involutionInMDI}
The following shows that MDI influence in Eq.~\ref{eq:mdi} is an involution.  Let $r_i = \operatorname{MDI}_\alpha(d_i)$:
\begin{equation}
\begin{split}
r_i & = (1 - d_i^\alpha)^{\frac{1}{\alpha}} \\
r_i^\alpha   & = 1 - d_i^\alpha \\
d_i^\alpha & = 1 - r_i^\alpha \\
d_i & = (1 - r_i^\alpha)^{\frac{1}{\alpha}}
\end{split}
\end{equation}

\subsection{Details of datasets}
\label{app:datasets}
Further details of the five datasets are in Table~\ref{tab:dataset-bins}.
\renewcommand{\arraystretch}{1.25}
\begin{table}[h]
  \caption{Dataset statistics and bin composition (sorted by imbalance ratio $\rho$)}
  \label{tab:dataset-bins}
  \centering\scriptsize
  \resizebox{\columnwidth}{!}{%
  \begin{tabular}{lrrrrrr}
    \toprule
    Dataset & Imb. $\rho$ & Lower Thr. & Upper Thr. & \% Bin 1 & \% Bin 2 & \% Bin 3 \\
    \midrule
    SEP-C   & \textit{1,476}  & --          & $\ln(10)$             & 98.04 & --    & 1.96 \\
    SARCOS  & \textit{3,267}  & $-0.5$      & $0.5$                 & 2.77  & 89.03 & 8.20 \\
    ONP     & \textit{6,746}  & $\log(350)$ & $\log(35\mathrm{k})$  & 0.81  & 98.30 & 0.89 \\
    SEP-EC  & \textit{10,478} & $-0.5$      & $0.5$                 & 0.77  & 98.09 & 1.14 \\
    BF      & \textit{33,559} & $\log(4)$   & $\log(40)$            & 80.81 & 15.86 & 3.33 \\
    \bottomrule
  \end{tabular}}
\end{table}

\subsection{Architecture and training hyperparameters}
\label{app:arch}
For each dataset, the MLP architecture is in Table~\ref{tab:arch} and the hyperparameters are in Table~\ref{tab:opt-hparams}.
\renewcommand{\arraystretch}{1.15}
\begin{table}[h]
  \caption{Dataset-specific MLP architecture}
  \label{tab:arch}
  \centering\scriptsize
  \begin{tabular}{lp{3.8cm}cc}
    \toprule
    Dataset & Hidden-layer widths & Embed dim & Dropout \\
    \midrule
    BF      & 4096–512–2048–512–1024–512 & 512 & 0.10 \\
    ONP     & 2048–128–1024–128–512–128–256–128 & 128 & 0.10 \\
    SARCOS  & 512–32–256–32–128–32–64–32 & 32  & 0.20 \\
    SEP-C   & 512–32–256–32–128–32–64–32 & 32  & 0.50 \\
    SEP-EC  & 2048–128–1024–128–512–128–256–128 & 128 & 0.20 \\
    \bottomrule
  \end{tabular}
\end{table}

\renewcommand{\arraystretch}{1.15}
\begin{table}[h]
  \caption{Training hyperparameters}
  \label{tab:opt-hparams}
  \centering\scriptsize
  \resizebox{\columnwidth}{!}{%
  \begin{tabular}{lrrrrr}
    \toprule
    Dataset & Batch size & Init. LR & W. decay & ES patience & Bandw. \\
    \midrule
    BF      & 4\,096  & $1\times10^{-4}$ & 0.01 & 3\,000 & 0.552 \\
    ONP     & 16\,384 & $5\times10^{-4}$ & 0.10 & 3\,000 & 1.429 \\
    SARCOS  & 14\,800 & $5\times10^{-4}$ & 0.10 & 3\,000 & 1.508 \\
    SEP-C   &   200   & $5\times10^{-4}$ & 1.00 & 3\,000 & 0.880 \\
    SEP-EC  & 2\,400  & $1\times10^{-4}$ & 0.10 & 4\,000 & 0.070 \\
    \bottomrule
  \end{tabular}}
\end{table}

\subsection{\texorpdfstring{Tuning hyperparameters $\alpha_e$, $\alpha_c$, and $\lambda$}{Tuning hyperparameters alphae, alphac, and lambda}}
\label{app:tuning}
In practice, we tune the three hyperparameters sequentially. 
We first fix $\lambda=0.5$ to start with, giving wPCC its default secondary role in the loss, and vary $\alpha_e$ to ensure adequate rare influence through wMSE. Then, keeping this $\alpha_e$ fixed, we vary $\lambda$ to jointly lower error and raise correlation. Only if the correlation on rare instances remains unsatisfactory do we set and vary $\alpha_c$ to give more influence to rare instances in wPCC.

\subsection{Weaker performance of SQINV+LDS+FDS and Balanced MSE}
\label{app:whyWeaker}
Our datasets are tabular, their datasets usually contain images.  We use the authors' implementations as well as their hyperparameters.  While their hyperparameters are effective for their datasets, they might not be as effective in our datasets. 

\subsection{MSE decomposition}
\label{app:mse-decomp}
\begin{align}
\mathrm{MSE}(\hat y,y)
&= (\bar y-\bar{\hat y})^{2}
   + \operatorname{var}(\hat y)
   + \operatorname{var}(y)
   - 2\,\operatorname{cov}(\hat y,y) \tag{1} \\[2pt]
&= (\bar y-\bar{\hat y})^{2}
   + \bigl(\operatorname{var}(\hat y)+\operatorname{var}(y)
           - 2\,\operatorname{sd}(\hat y)\operatorname{sd}(y)\bigr) \notag\\
&\quad + 2\,\operatorname{sd}(\hat y)\operatorname{sd}(y)
   - 2\,\operatorname{cov}(\hat y,y) \tag{2} \\[2pt]
&= (\bar y-\bar{\hat y})^{2}
   + \bigl(\operatorname{sd}(\hat y)-\operatorname{sd}(y)\bigr)^{2} \notag\\
&\quad + 2\bigl(\operatorname{sd}(\hat y)\operatorname{sd}(y)
           - \operatorname{cov}(\hat y,y)\bigr) \tag{3} \\[2pt]
\mathrm{MSE}(\hat y,y)
&= (\bar y-\bar{\hat y})^{2}
   + \bigl(\operatorname{sd}(\hat y)-\operatorname{sd}(y)\bigr)^{2} \notag\\
&\quad + 2\,\operatorname{sd}(\hat y)\operatorname{sd}(y)\,
     \bigl(1-\operatorname{PCC}(\hat y,y)\bigr). \tag{4}
\end{align}

\end{document}